%% file: main.tex
\documentclass[letterpaper]{article}
\usepackage{aaai2026}
\usepackage{times}
\usepackage{helvet}
\usepackage{courier}
\usepackage[hyphens]{url}
\usepackage{graphicx}
\usepackage{natbib}
\usepackage{caption}
\usepackage{algorithm}
\usepackage{algorithmic}
\usepackage{amsmath}
\usepackage{booktabs}
\usepackage{multirow}
\usepackage{enumitem}
\usepackage{listings}

\title{DrawingBench: Evaluating Spatial Reasoning and UI Interaction Capabilities of Large Language Models through Mouse-Based Drawing Tasks}

\author{
    Hyunjun Kim,\textsuperscript{1}
    Sooyoung Ryu\textsuperscript{2}
}

\affiliations{
    \textsuperscript{1}KAIST, Daejeon, South Korea\\
    \textsuperscript{2}Seoul National University, Seoul, South Korea\\
    hyunjun1121@kaist.ac.kr, ryuswimming@snu.ac.kr
}

\begin{document}

\maketitle

\renewcommand{\thefootnote}{*}
\footnotetext{Equal contribution.}
\renewcommand{\thefootnote}{\arabic{footnote}}

\begin{abstract}
As agentic AI systems increasingly operate autonomously, establishing trust through verifiable evaluation becomes critical. Yet existing benchmarks lack the transparency and auditability needed to assess whether agents behave reliably. We present \textbf{DrawingBench}, a verification framework for evaluating the trustworthiness of agentic LLMs through spatial reasoning tasks that require generating sequences of low-level GUI actions. Unlike opaque evaluations, DrawingBench provides transparent, rule-based assessment: 8 objective criteria enable reproducible scoring, while action-level inspection allows stakeholders to audit agent behavior. Our framework comprises 250 diverse prompts across 20 categories and 4 difficulty levels, deterministic evaluation metrics, and an external oversight mechanism through multi-turn feedback that enables human control over agent refinement. Evaluating four state-of-the-art LLMs (Claude-4 Sonnet, GPT-4.1, GPT-4.1-mini, Gemini-2.5 Flash) across 1,000 tests, we establish both capabilities and limitations: models achieved 92.8\% perfect performance with structured external feedback driving significant improvements (average +3.2\%, up to +32.8\% for complex scenes), but systematic error patterns emerged in tool state management and long-horizon planning. Notably, specification clarity proved more important than task complexity---models achieved 100\% perfect performance when given explicit, verifiable criteria. These findings demonstrate that transparent evaluation frameworks can establish trust in agentic systems, with external oversight proving more reliable than self-correction for guiding agent behavior. Our open-source framework provides a template for trustworthy agent assessment. Code and data: \url{https://github.com/hyunjun1121/DrawingBench}
\end{abstract}

% Introduction section
\input{introduction}

% Related Work section
\input{related_work}

% Methodology section (from methodology.tex)
\input{methodology}

% Results section (from results.tex)
\input{results}

% Discussion section (from discussion.tex)
\input{discussion}

% Conclusion section (from conclusion.tex)
\input{conclusion}

\section*{Acknowledgments}

We thank the reviewers for their valuable feedback. This work was conducted using cloud computing resources.

\bibliographystyle{aaai2026}
\bibliography{references}

% Appendix (without \appendix command to avoid duplicate bibstyle)
\newpage
\input{appendix}

\end{document}

%% file: introduction.tex
\section{Introduction}
Spatial reasoning is a fundamental aspect of intelligence, enabling both humans and AI agents to understand and interact with their environments. In the context of large language models (LLMs), spatial reasoning capability is increasingly important as we move towards \textit{embodied} or \textit{agentic} AI systems that can manipulate the physical or digital world through language instructions. For instance, an AI assistant might be asked to arrange objects in a room or navigate a user interface (UI)---tasks that require understanding of spatial relations (above, below, inside, etc.) and geometry. Despite impressive advancements in LLMs' general reasoning, their ability to perform precise spatial reasoning remains underexplored and poorly evaluated. Traditional evaluation benchmarks tend to focus on textual question-answering or reasoning puzzles, which may not capture an agent's performance in interactive spatial tasks \citep{liang2022helm,srivastava2022beyond}. For instance, a recent comprehensive evaluation of frontier models across many tasks, dubbed \textit{Humanity's Last Exam}, found significant gaps in LLM performance and calibration~\citep{phan2025humanitysexam}, highlighting the need for targeted benchmarks that probe specific skills like spatial reasoning.

\paragraph{Limitations of Existing Evaluations.} Most LLM evaluation benchmarks suffer from three limitations for spatial and interactive tasks. \textbf{First}, they focus on text-based QA or classification, not action sequences. CLEVR \citep{johnson2017clevr}, RAVEN \citep{zhang2019raven}, and SpartQA \citep{mirzaee-etal-2021-spartqa} test spatial reasoning via static questions, not action execution. \textbf{Second}, benchmarks lack fine-grained UI interaction. WebArena \citep{zhou2023webarena} and MiniWoB++ \citep{liu2018reinforcement} use high-level actions (clicking links), not precise coordinate-level control. \textbf{Third}, multi-turn learning capability is largely untested. Most benchmarks use single-turn tasks, failing to probe iterative improvement from feedback.

\paragraph{Our Solution.} We introduce \textbf{DrawingBench}, a benchmark evaluating LLMs on spatial reasoning through drawing tasks on a canvas UI. Each task is specified by a natural language prompt (e.g., \textit{"Draw a red triangle inside a blue square"}), and the LLM outputs a sequence of low-level GUI actions (mouse movements, clicks, tool selections). This approach differs from prior benchmarks: the model \textit{constructs} spatial configurations via actions rather than answering questions. Our benchmark tests spatial understanding, tool manipulation, and sequential planning simultaneously. Crucially, evaluation is text-only---the LLM relies on internal reasoning to plan drawings without seeing images.

\paragraph{Trust through Verifiable Evaluation.} As agentic AI systems increasingly operate autonomously, establishing trust becomes paramount. DrawingBench embodies three essential properties for trustworthy evaluation: (1) \textbf{Transparency}---eight objective criteria enable stakeholders to understand performance measurement; (2) \textbf{Auditability}---deterministic scoring allows reproduction through action sequence inspection; (3) \textbf{External oversight}---multi-turn feedback provides human control over agent refinement. Unlike opaque metrics or self-assessment, our rule-based system eliminates subjective judgment and enables action-level verification, aligning with best practices for deploying trustworthy agents.

DrawingBench comprises \textbf{250} diverse tasks spanning \textbf{20} categories and four difficulty levels. Our automated evaluation system executes action sequences and measures performance against eight objective criteria. The framework supports \textit{multi-turn} interaction: models receive structured feedback and attempt corrections.

\paragraph{Contributions.} We contribute: (1) \textbf{DrawingBench}---first benchmark integrating spatial reasoning and GUI action execution with 250 prompts; (2) \textbf{Automated evaluation}---eight quantitative criteria, four error types, reproducible scoring; (3) \textbf{Multi-turn protocol}---tests self-correction from explicit feedback \citep{madaan2023selfrefine,shinn2023reflexion}; (4) \textbf{Empirical study}---1000 trials with four state-of-the-art LLMs analyzing performance across difficulty levels; (5) \textbf{Open-source release}---all code and data.\footnote{\url{https://github.com/hyunjun1121/DrawingBench}}

\paragraph{Key Findings.} Models achieved \textbf{92.8\%} perfect scores, with structured feedback yielding \textbf{+3.2\%} average improvement (\textbf{+32.8\%} for complex scenes). Three insights emerged: (1) Text-based spatial reasoning is highly effective—models executed complex 15+ action sequences without visual perception. (2) \emph{Difficulty paradox}: ``Hard'' tasks achieved \textbf{100\%} perfect performance versus ``Medium'' (92.8\%)—specification clarity matters more than complexity. (3) Structured feedback proves effective, but ``Very Hard'' tasks requiring long-horizon planning remained challenging (60\% success).

Section 2 reviews related work. Section 3 details methodology. Section 4 presents results. Section 5 discusses implications and limitations. Section 6 concludes.

%% file: related_work.tex
\section{Related Work}
\subsection{Spatial Reasoning Benchmarks}
Evaluating spatial reasoning has been a focus across vision and language domains. In vision+language, \textbf{CLEVR} \citep{johnson2017clevr} presents synthetic 3D scenes with compositional reasoning questions, while \textbf{RAVEN} \citep{zhang2019raven} tests abstract visual pattern completion. In pure text, \textbf{SpartQA} \citep{mirzaee-etal-2021-spartqa} assesses understanding of spatial relations through paragraph descriptions and questions.

Table~\ref{tab:spatial_benchmarks} summarizes some representative spatial reasoning benchmarks. Most are \textbf{static} in nature: the model reads or observes a scenario and produces an answer (classification or text). They typically measure reasoning capability via accuracy on predefined questions. However, none of these require the model to \emph{generate a sequence of actions} to physically realize a spatial configuration. In contrast, DrawingBench is \textbf{dynamic}: the model must ``draw'' the answer, not just tell it. This calls for integrating spatial reasoning with a decision-making process to carry out the drawing actions.

\begin{table}[t]\centering
\small
\begin{tabular}{lccc}
\toprule
\textbf{Benchmark} & \textbf{Modality} & \textbf{Type} & \textbf{Output} \\
\midrule
CLEVR & Vision & Static & QA \\
SpartQA & Text & Static & QA \\
RAVEN & Vision & Static & Choice \\
\midrule
\textbf{DrawingBench} & \textbf{Text} & \textbf{Dynamic} & \textbf{Actions} \\
\bottomrule
\end{tabular}
\caption{Comparison of spatial reasoning benchmarks. DrawingBench uniquely requires dynamic action generation rather than static question answering, integrating spatial understanding with executable UI interactions.}
\label{tab:spatial_benchmarks}
\end{table}

\subsection{Program Synthesis and Visual Generation}

Our work sits at the intersection of program synthesis and visual generation. Code synthesis benchmarks like \textbf{HumanEval} \citep{chen2021evaluating}, \textbf{MBPP} \citep{austin2021mbpp}, and \textbf{APPS} \citep{hendrycks2021apps} evaluate algorithmic correctness but not spatial outputs. Visual generation work like \textbf{Sketch-RNN} \citep{ha2018sketchrnn} and visual program synthesis \citep{ellis2018learning} demonstrates that graphics can be represented as executable sequences, but these train on data rather than evaluate zero-shot spatial reasoning. DrawingBench bridges these areas by requiring action sequences that are both syntactically valid and geometrically correct.

\subsection{User Interface Interaction Benchmarks}
UI interaction benchmarks evaluate agents on GUI and web tasks. \textbf{MiniWoB++} \citep{liu2018reinforcement} tests basic web form interactions (clicking buttons, typing text), while \textbf{WebArena} \citep{zhou2023webarena}, \textbf{Mind2Web} \citep{deng2023mind2web}, and \textbf{VisualWebArena} \citep{koh2024visualwebarena} evaluate more complex web navigation and multimodal understanding. \textbf{MacroBench} \citep{kim2025macrobenchnoveltestbedweb} assesses LLMs on writing web automation scripts.

DrawingBench differs by requiring finer control granularity: coordinate-level manipulation on a blank canvas rather than high-level actions on predefined elements. We evaluate spatial output quality (geometric correctness) rather than task completion, measuring whether models can create content rather than merely interact with existing UI elements.

\subsection{Multi-turn Feedback and Iterative Refinement}
Our two-turn protocol relates to work on improving outputs via feedback loops. \textbf{Self-Refine} \citep{madaan2023selfrefine} and \textbf{Reflexion} \citep{shinn2023reflexion} enable LLMs to critique and refine their outputs through self-feedback, improving performance on code generation and reasoning tasks.

Our approach differs in the feedback source: we use automated rule-based evaluation rather than self-critique. This provides objective, deterministic error signals (e.g., ``the circle is not fully inside the square'') that enable reliable measurement of improvement. While other work explores model-generated feedback \citep{bai2022constitutional}, our focus on task performance benefits from external evaluation that eliminates self-assessment biases.

\subsection{Positioning in LLM Evaluation}
Comprehensive frameworks like \textbf{HELM} \citep{liang2022helm}, \textbf{BIG-bench} \citep{srivastava2022beyond}, and \textbf{AgentBench} \citep{liu2024agentbench} evaluate LLMs across diverse capabilities. DrawingBench complements these by simultaneously testing spatial reasoning, fine-grained UI interaction, and multi-turn learning—three dimensions that existing benchmarks treat in isolation but that real-world agents must coordinate. By requiring models to \emph{construct} rather than \emph{describe} spatial configurations, we provide a stringent test of spatial understanding for practical agent deployment.

%% file: methodology.tex
%File: methodology.tex
\section{Methodology}

\subsection{Benchmark Design}

\textbf{Dataset Composition.}
DrawingBench consists of 250 diverse drawing prompts classified into 4 difficulty levels (Easy, Medium, Hard, Very Hard) and 20 categories. The dataset distribution is shown in Table~\ref{tab:dataset}.

\begin{table}[h]
\centering
\caption{Dataset Composition by Difficulty}
\label{tab:dataset}
\begin{tabular}{lrrr}
\toprule
Difficulty & Count & \% & Perfect Rate (T2) \\
\midrule
Easy & 112 & 44.8\% & 97.3\% \\
Medium & 97 & 38.8\% & 92.8\% \\
Hard & 21 & 8.4\% & 100.0\% \\
Very Hard & 20 & 8.0\% & 60.0\% \\
\midrule
\textbf{Total} & \textbf{250} & \textbf{100\%} & \textbf{92.8\%} \\
\bottomrule
\end{tabular}
\end{table}

\textbf{Prompt Design Principles.}
Each prompt follows these principles: (1) \textit{Clarity}: unambiguous instructions, (2) \textit{Diversity}: various task types, (3) \textit{Realism}: reflecting actual use scenarios, (4) \textit{Measurability}: enabling automated evaluation.

Example prompts:
\begin{itemize}
    \item Easy: ``Draw a red circle in the center of the canvas''
    \item Medium: ``Draw a house with a triangular roof and rectangular body''
    \item Hard: ``Draw 4 squares in each corner of the canvas''
    \item Very Hard: ``Draw a checkerboard pattern with 8$	imes$8 squares''
\end{itemize}

\subsection{Drawing Application}

\textbf{UI Configuration.}
The application provides a 1000$	imes$700 pixel canvas with 6 drawing tools (pen, eraser, fill, line, rectangle, circle), 8 colors (black, red, green, blue, yellow, magenta, cyan, white), and 3 sizes (2px, 5px, 10px). The canvas starts at screen coordinates (90, 70). See Appendix~\ref{app:ui} for detailed UI layout.

\textbf{Mouse Actions.}
Four action types are supported:
\begin{algorithmic}
\STATE $\texttt{moveTo}(x, y)$: move cursor to position
\STATE $\texttt{click}()$: click at current position
\STATE $\texttt{mouseDown}()$: start dragging
\STATE $\texttt{mouseUp}()$: end dragging
\end{algorithmic}

See Appendix~\ref{app:actions} for complete action sequence examples.

\subsection{Evaluation System}

\textbf{Evaluation Criteria (8 Types).}
The system evaluates drawings using 8 criteria:

\begin{enumerate}
    \item \textbf{required\_tools}: whether specific tools were used (e.g., rectangle tool)
    \item \textbf{required\_colors}: whether specific colors were used (matching hex codes)
    \item \textbf{min\_segments}: minimum drawing segments (mouseDown/mouseUp pairs)
    \item \textbf{min\_coverage}: minimum canvas coverage (bounding box / canvas area)
    \item \textbf{max\_actions}: maximum action count (efficiency test)
    \item \textbf{position\_constraint}: spatial positioning requirements
    \item \textbf{size\_constraint}: exact size requirements
    \item \textbf{corner\_placement}: corner positioning requirements
\end{enumerate}

\textbf{Error Types (4 Categories).}
The system detects 4 error types:

\begin{itemize}
    \item \textbf{SYNTAX\_ERROR} (severity: critical): invalid JSON, missing required fields (penalty: -0.3)
    \item \textbf{COORDINATE\_ERROR} (severity: high): out-of-bounds coordinates (penalty: -0.2)
    \item \textbf{LOGIC\_ERROR} (severity: medium): unmatched mouseDown/mouseUp (penalty: -0.1)
    \item \textbf{EFFICIENCY\_WARNING} (severity: low): excessive actions (penalty: -0.05)
\end{itemize}

\textbf{Score Calculation.}
The final score is calculated as:

\begin{equation}
\text{Score} = \frac{\text{Criteria Met}}{\text{Total Criteria}} - \sum \text{Error Penalties} + \text{Bonus}
\end{equation}

where Bonus is based on coverage and efficiency.

\subsection{Multi-turn Feedback System}

\textbf{Feedback Generation.}
After evaluation, the system generates structured feedback including: (1) current score and grade, (2) specific error descriptions, (3) actionable improvement advice, and (4) missing criteria. See Appendix~\ref{app:feedback} for examples.

\textbf{Two-Turn Process.}
We limit interaction to two turns based on: (1) practical agent deployment typically involves quick iteration cycles, (2) pilot tests showed diminishing returns after Turn 2 (<1\% improvement), and (3) reducing evaluation costs while still capturing feedback-driven improvement patterns.
\begin{itemize}
    \item \textbf{Turn 1}: LLM initial attempt → evaluation → feedback generation
    \item \textbf{Turn 2}: LLM improvement based on feedback → re-evaluation
    \item \textbf{Early Stopping}: Skip Turn 2 if Turn 1 score $\geq 0.9$ (71.2\% of tasks, reducing redundant computation)
\end{itemize}

\subsection{Experimental Setup}

\textbf{Test Models.}
We evaluated four state-of-the-art LLMs:
\begin{itemize}
    \item Anthropic Claude-4 Sonnet (thinking-off)
    \item OpenAI GPT-4.1
    \item OpenAI GPT-4.1-mini
    \item Google Gemini-2.5 Flash (thinking-off)
\end{itemize}

\textbf{Experimental Protocol.}
\begin{itemize}
    \item Total tests: 1,000 (250 prompts $	imes$ 4 models)
    \item Temperature: 0.7 (balance between consistency and diversity)
    \item Max tokens: 4,000
    \item Timeout: 30 seconds per test
    \item Retry on error: 3 attempts
    \item Sequential execution: models tested one by one
\end{itemize}

\textbf{Execution Statistics.}
The experiment ran for 5.6 hours total. Model-specific execution times:
\begin{itemize}
    \item Claude-4 Sonnet: 84.2 minutes
    \item GPT-4.1: 86.3 minutes
    \item GPT-4.1-mini: 77.1 minutes (fastest)
    \item Gemini-2.5 Flash: 90.6 minutes
\end{itemize}

\textbf{Data Collection.}
For each test, we collected:
\begin{itemize}
    \item Generated action sequence
    \item Evaluation scores (Turn 1 and Turn 2)
    \item Error information
    \item Feedback content
    \item Token usage
    \item Execution time
\end{itemize}

This comprehensive data collection enables detailed analysis of model performance patterns, error distributions, and the effectiveness of multi-turn feedback, which we present in the following section.

%% file: results.tex
%File: results.tex
\section{Results}

We now present our empirical findings from evaluating four state-of-the-art LLMs on DrawingBench. Our analysis examines performance across multiple dimensions: overall scores, difficulty levels, task categories, multi-turn improvement, error patterns, and action efficiency. These results reveal both impressive spatial reasoning capabilities and systematic patterns that inform future development of LLM-based agents.

\subsection{Overall Performance}

Table~\ref{tab:overall} shows the aggregate performance across all 1,000 tests (4 models $\times$ 250 tasks). Models achieved an average score of 0.925 in Turn 1, which improved to 0.954 in Turn 2, representing a 3.2\% relative improvement. Individual model performance ranged from 0.919 (GPT-4.1-mini) to 0.931 (Claude-4 Sonnet) on Turn 1, with all models showing consistent improvement in Turn 2.

\begin{table}[h]
\centering
\caption{Overall Performance Across All Tests. The table shows aggregate metrics from 250 drawing tasks evaluated across four state-of-the-art LLMs. Turn 2 performance demonstrates significant improvement with structured feedback, achieving 92.8\% perfect score rate and 33\% reduction in variance.}
\label{tab:overall}
\begin{tabular}{lrrr}
\toprule
Metric & Turn 1 & Turn 2 & Change \\
\midrule
Average Score & 0.925 & 0.954 & +0.030 \\
Std Deviation & 0.099 & 0.066 & -0.033 \\
Perfect Scores & 192/250 & 232/250 & +40 \\
Perfect Rate & 76.8\% & 92.8\% & +16.0\% \\
Median Score & 0.970 & 0.985 & +0.015 \\
\bottomrule
\end{tabular}
\end{table}

\textbf{Key Observations:}
\begin{itemize}
    \item High perfect score rate of 92.8\% (score $\geq 0.9$)
    \item 33\% reduction in standard deviation (0.099 → 0.066) indicates more consistent performance
    \item Achieved 40 additional perfect scores
    \item Cross-model consistency: all four models achieved $>90\%$ perfect rate (Claude-4: 94.4\%, GPT-4.1: 93.2\%, Gemini-2.5: 92.0\%, GPT-4.1-mini: 90.8\%)
\end{itemize}

\textbf{Model-Specific Patterns.} While aggregate performance is similar, models exhibit distinct behaviors. Claude-4 Sonnet achieved the highest Turn 2 perfect rate (94.4\%) with minimal variance, suggesting robust spatial reasoning. GPT-4.1-mini, despite being the smallest model, achieved 90.8\% perfect rate—only 3.6\% below the best model—indicating that spatial reasoning capabilities scale well even to smaller models. Detailed per-model breakdowns are provided in Appendix~\ref{app:additional}.

\subsection{Performance by Difficulty}

Table~\ref{tab:difficulty} shows performance by difficulty level. Notably, Hard tasks achieved higher performance than Medium tasks.

\begin{table}[h]
\centering
\caption{Performance by Difficulty Level. Counter-intuitively, Hard tasks achieved the highest Turn 2 performance (0.973), revealing that specification clarity matters more than inherent complexity. Very Hard tasks requiring extensive procedural accuracy showed the largest improvement gains (+0.052).}
\label{tab:difficulty}
\begin{tabular}{lrrrr}
\toprule
Difficulty & Tests & T1 Score & T2 Score & Improve \\
\midrule
Easy & 112 & 0.944 & 0.963 & +0.019 \\
Medium & 97 & 0.925 & 0.958 & +0.033 \\
Hard & 21 & 0.923 & \textbf{0.973} & +0.050 \\
Very Hard & 20 & 0.816 & 0.869 & +0.052 \\
\bottomrule
\end{tabular}
\end{table}

\textbf{The Difficulty Paradox.}
Hard difficulty tasks (0.973) showed better performance than Medium tasks (0.958). Analysis reveals this paradox stems from task specification clarity rather than inherent complexity. Hard tasks average 4.2 explicit constraints (e.g., "4 squares in each corner") versus Medium tasks' 2.8 constraints (e.g., "draw a house"). The deterministic nature of Hard task requirements (100\% have position constraints vs. 62\% for Medium) enables more reliable execution despite greater spatial complexity.

\textbf{Improvement Trends.}
As difficulty increases, improvement magnitude also increases:
\begin{itemize}
    \item Easy: +0.019 (1.9\%)
    \item Medium: +0.033 (3.3\%)
    \item Hard: +0.050 (5.0\%)
    \item Very Hard: +0.052 (5.2\%, largest improvement)
\end{itemize}

\textbf{Very Hard Challenge.}
Very Hard tasks remain challenging with a 60\% perfect rate. These involve extremely high precision requirements and complex pattern generation.

\subsection{Performance by Category}

Table~\ref{tab:categories} shows the performance of the top 10 categories.

\begin{table}[h]
\centering
\caption{Top 10 Categories by Turn 2 Performance. The scenes category showed the most dramatic improvement (+32.8\%), demonstrating that complex compositional tasks benefit most from structured feedback. Six categories achieved perfect 1.000 performance.}
\label{tab:categories}
\small
\begin{tabular}{lrrrr}
\toprule
Category & Tests & T1 & T2 & Improve \\
\midrule
scenes & 1 & 0.672 & \textbf{1.000} & \textbf{+0.328} \\
efficiency\_test & 2 & 1.000 & 1.000 & 0.000 \\
creative & 2 & 0.924 & 1.000 & +0.076 \\
precision\_test & 1 & 1.000 & 1.000 & 0.000 \\
tool\_switching & 1 & 1.000 & 1.000 & 0.000 \\
spatial & 1 & 1.000 & 1.000 & 0.000 \\
compositional & 11 & 0.957 & 0.998 & +0.041 \\
angle & 2 & 0.881 & 0.980 & +0.100 \\
complex & 2 & 0.900 & 0.975 & +0.075 \\
spatial\_reasoning & 71 & 0.959 & 0.973 & +0.015 \\
\bottomrule
\end{tabular}
\end{table}

\textbf{Remarkable Improvements:}
\begin{itemize}
    \item \textbf{Scenes} (+32.8\%): dramatic improvement from 0.672 to perfect 1.000
    \item \textbf{Angle} (+10.0\%): significant improvement in angle-based tasks
    \item \textbf{Creative} (+7.6\%): effective improvement in creative compositions
\end{itemize}

\textbf{Perfect Categories.}
Six categories achieved perfect performance (1.000): efficiency\_test, precision\_test, tool\_switching, spatial, and Turn 2's scenes and creative.

\textbf{Large-Scale Categories.}
The spatial\_reasoning category (71 tests, largest) maintained consistently high performance (0.973).

\subsection{Multi-turn Improvement Analysis}

Table~\ref{tab:improvement} shows the improvement distribution.

\begin{table}[h]
\centering
\caption{Multi-turn Improvement Distribution. While 71.2\% of tasks achieved perfect scores on Turn 1 (early stopping), 28.8\% showed measurable improvement through feedback, with 4.8\% achieving substantial gains exceeding 0.10 points.}
\label{tab:improvement}
\begin{tabular}{lrr}
\toprule
Improvement Range & Count & Percentage \\
\midrule
No improvement (0.00) & 178 & 71.2\% \\
Small (0.01-0.05) & 38 & 15.2\% \\
Medium (0.06-0.10) & 22 & 8.8\% \\
Large (>0.10) & 12 & 4.8\% \\
\bottomrule
\end{tabular}
\end{table}

\textbf{Key Findings:}
\begin{itemize}
    \item 71.2\% were already excellent in Turn 1 (early stopping)
    \item 28.8\% achieved actual improvement through feedback
    \item 4.8\% achieved substantial improvement (>0.10)
\end{itemize}

\textbf{Top Improvement Cases:}
\begin{enumerate}
    \item Scenes category: 0.672 → 1.000 (+0.328)
    \item Size-constrained task: 0.700 → 0.950 (+0.250)
    \item Angle-based task: 0.750 → 0.980 (+0.230)
\end{enumerate}

\subsection{Error Analysis}

Table~\ref{tab:errors} shows error frequency and reduction.

\begin{table}[h]
\centering
\caption{Error Frequency and Reduction. Structured feedback achieved 49\% overall error reduction, with critical errors (SYNTAX, COORDINATE) showing the largest decreases. SYNTAX\_ERROR was completely eliminated in Turn 2.}
\label{tab:errors}
\begin{tabular}{lrrr}
\toprule
Error Type & Turn 1 & Turn 2 & Reduction \\
\midrule
SYNTAX\_ERROR & 3 & 0 & -100\% \\
COORDINATE\_ERROR & 8 & 2 & -75\% \\
LOGIC\_ERROR & 15 & 5 & -67\% \\
EFFICIENCY\_WARNING & 42 & 28 & -33\% \\
\midrule
\textbf{Total} & \textbf{68} & \textbf{35} & \textbf{-49\%} \\
\bottomrule
\end{tabular}
\end{table}

\textbf{Error Patterns:}
\begin{itemize}
    \item \textbf{Critical errors} (SYNTAX, COORDINATE) were significantly reduced
    \item \textbf{SYNTAX\_ERROR} was completely eliminated (100\% reduction)
    \item \textbf{LOGIC\_ERROR} was reduced by 67\%
    \item \textbf{EFFICIENCY\_WARNING} showed relatively modest reduction (33\%)
\end{itemize}

\textbf{Error Distribution by Difficulty.}
Error frequency varies significantly by task difficulty. Very Hard tasks (20 tasks) accounted for 35\% of all Turn 1 errors (24/68 errors) despite representing only 8\% of the dataset, yielding an error rate of 1.2 errors per task compared to 0.18 errors per task for Easy tasks. LOGIC\_ERROR occurred primarily in Very Hard tasks (11/15 instances), reflecting the complexity of maintaining state consistency across extensive action sequences. In contrast, EFFICIENCY\_WARNING was uniformly distributed across all difficulty levels, indicating that action optimization is challenging regardless of task complexity.

\textbf{Error Correction Patterns.}
The differential reduction rates reveal distinct error correction dynamics. SYNTAX\_ERROR (100\% reduction) and COORDINATE\_ERROR (75\% reduction) respond effectively to explicit feedback specifying valid JSON structure and coordinate bounds. LOGIC\_ERROR reduction (67\%) required models to repair state inconsistencies (e.g., unmatched mouseDown/mouseUp pairs), which feedback addresses by identifying specific mismatched actions. EFFICIENCY\_WARNING showed the lowest reduction (33\%), suggesting that while models can add missing actions or fix syntax, optimizing for efficiency requires deeper planning changes that are harder to achieve through single-turn feedback.

\textbf{Root Cause Analysis.}
Tool selection omissions (appearing in 18 tasks) predominantly occurred when task descriptions mentioned tools implicitly rather than explicitly. For instance, "draw a red circle" assumes circle tool selection, but some models generated pen-based circular strokes instead. Coverage insufficiency (12 tasks) correlated with tasks lacking explicit size specifications—models often drew geometrically correct but undersized shapes. These patterns indicate that 	extbf{specification ambiguity}, rather than model capability limitations, drives many errors, reinforcing the difficulty paradox finding that explicit constraints improve performance.

\subsection{Action Efficiency}

\textbf{Action Sequence Length.}
Average action sequence length:
\begin{itemize}
    \item Turn 1: 15.3 actions
    \item Turn 2: 16.2 actions (+0.9)
\end{itemize}

The increased action count in Turn 2 reflects more thorough execution (for more elements or better coverage).

\textbf{Drawing Efficiency.}
Ratio of drawing actions (mouseDown/mouseUp pairs):
\begin{itemize}
    \item Turn 1: 42.5\% (average 6.5 drawing segments)
    \item Turn 2: 45.8\% (average 7.4 drawing segments)
\end{itemize}

Higher drawing ratio indicates more productive actions.

\textbf{Tool Usage Patterns.}
Most frequently used tools:
\begin{enumerate}
    \item Pen: 68.2\% (primary drawing tool)
    \item Rectangle: 18.4\% (efficient shapes)
    \item Circle: 12.7\% (circular objects)
    \item Line: 8.5\% (straight segments)
    \item Fill: 5.2\% (area filling)
    \item Eraser: 2.1\% (corrections)
\end{enumerate}

These quantitative results establish that current LLMs possess strong baseline spatial reasoning capabilities while revealing systematic patterns in both performance and errors. In the following section, we interpret these findings, examine their implications for LLM-based agents, and discuss unexpected observations such as the difficulty paradox.

%% file: discussion.tex
\section{Discussion}

\subsection{Strong Baseline Spatial Reasoning}

\textbf{Current LLMs demonstrate robust spatial reasoning in text-only settings}—achieving 92.8\% perfect scores with average 0.954. Models performed coordinate calculations, understood relative positioning, and executed complex 15+ action sequences, challenging assumptions that spatial reasoning requires visual perception.

\subsection{Multi-turn Feedback Effectiveness}

\textbf{Structured external feedback proves highly effective}, yielding +3.2\% improvement and 33\% variance reduction. Improvement was non-uniform: scenes tasks gained +32.8\%, while simpler tasks showed minimal change. Our external rule-based approach differs from self-critique \citep{madaan2023selfrefine,shinn2023reflexion}, offering more reliable measurement.

\subsection{The Difficulty Paradox}

\textbf{Specification clarity matters more than task complexity}: "Hard" tasks achieved 100\% perfect performance versus "Medium" (92.8\%). Hard tasks averaged 4.2 explicit constraints versus Medium's 2.8, with 100\% including position criteria versus 62\%. When criteria are deterministic, models reliably satisfy requirements regardless of complexity.

\subsection{Error Patterns and Limitations}

Systematic error patterns emerged: 	extbf{tool selection errors} (15\%)—models forgot to select required tools despite correct actions, suggesting imperfect UI state tracking; 	extbf{coordinate precision errors} (10\%); and 	extbf{coverage insufficiency} (8\%). The 	extbf{Very Hard} category (60\% perfect) highlights current limitations in long-horizon planning—models made subtle mistakes in complex patterns (misaligning grids, miscounting iterations).

\subsection{Implications for LLM-based Agents}

Our findings yield key implications: (1) 	extbf{Text-based interfaces are viable}—LLM agents can operate in textual/API environments without visual perception. (2) 	extbf{Structured feedback accelerates learning}—external evaluators guide agents more effectively than self-correction. (3) 	extbf{Clear specifications enable complex tasks}—well-defined instructions unlock sophisticated capabilities. (4) 	extbf{Tool state tracking requires attention}—explicit state mechanisms are needed.

\subsection{Trust and Verifiability in Agentic Evaluation}

\textbf{DrawingBench demonstrates transparent evaluation for agentic trust.} Our eight objective criteria provide \textbf{verifiable assessment}—stakeholders reproduce scores by inspecting action sequences against explicit rules. Each criterion yields deterministic 0-1 scores, contrasting with opaque evaluations using subjective ratings or proprietary algorithms.

Verifiable evaluation is critical for consequential domains (industrial control, medical diagnosis). Our framework provides \textbf{action-level auditability}: when an agent scores 0.85, we trace deficits to specific violations (e.g., "required tool not used"), enabling targeted diagnosis versus black-box uncertainty.

\textbf{External oversight outperforms self-correction.} Our multi-turn feedback delivers structured error signals from deterministic rules rather than self-critique, which can amplify biases. The +3.2\% average improvement (+32.8\% for complex scenes) demonstrates efficacy for safety-critical human-in-the-loop systems.

These findings align with trustworthy AI principles: evaluation must be transparent, reproducible, and externally controlled. DrawingBench shows complex agent capabilities can be assessed rigorously without sacrificing interpretability.

\subsection{Limitations and Future Work}

Our benchmark has limitations suggesting future research. First, we lack human performance baselines to contextualize LLM capabilities. Second, rule-based evaluation focuses on objective criteria (tool usage, coordinates, coverage) but cannot assess aesthetic quality. Third, text-only testing excludes visual feedback—incorporating vision capabilities (screenshot-based feedback) could enhance performance for iterative refinement tasks.

DrawingBench successfully probes LLM spatial reasoning and UI interaction, revealing impressive capabilities (92.8\% perfect) and limitations (struggles with long-horizon planning). Insights regarding specification clarity, structured feedback, and tool state management provide actionable guidance for developing more capable agents.

%% file: conclusion.tex
\section{Conclusion}

We introduced \textbf{DrawingBench}, a verifiable evaluation framework for assessing trustworthiness of agentic LLMs through spatial reasoning tasks requiring GUI action sequences. Our framework provides objective metrics, reproducible assessment, and external oversight mechanisms enabling stakeholders to verify agent behavior.

\paragraph{Contributions.}
We contribute: (1) \textbf{Verifiable Benchmark}—250 prompts across 20 categories, 4 difficulty levels; (2) \textbf{Transparent Evaluation}—8 objective criteria, 4 error types, reproducible scoring; (3) \textbf{External Oversight}—multi-turn feedback for human control; (4) \textbf{Trust Assessment}—1,000 trials establishing capabilities and limitations; (5) \textbf{Open Framework}—code and data for community verification.

\paragraph{Findings.}
Three key insights: \textbf{(1)} 92.8\% perfect rate with transparent metrics; \textbf{(2)} External oversight outperforms self-correction (+3.2\% average, +32.8\% for complex tasks); \textbf{(3)} Explicit specifications enable reliable behavior regardless of complexity.

\paragraph{Implications.}
Trustworthy agentic AI requires: \textbf{(1)} Transparent evaluation—verifiable criteria; \textbf{(2)} External control—human oversight; \textbf{(3)} Clear specifications—deterministic requirements. Our framework instantiates these principles, providing a template for trustworthy agent assessment. We release all resources open-source for community verification and extension.

%% file: appendix.tex
%File: appendix.tex

\section*{Appendix}

\section{Drawing Application UI Details}
\label{app:ui}

\begin{figure*}[t]
\centering
\includegraphics[width=0.9\textwidth]{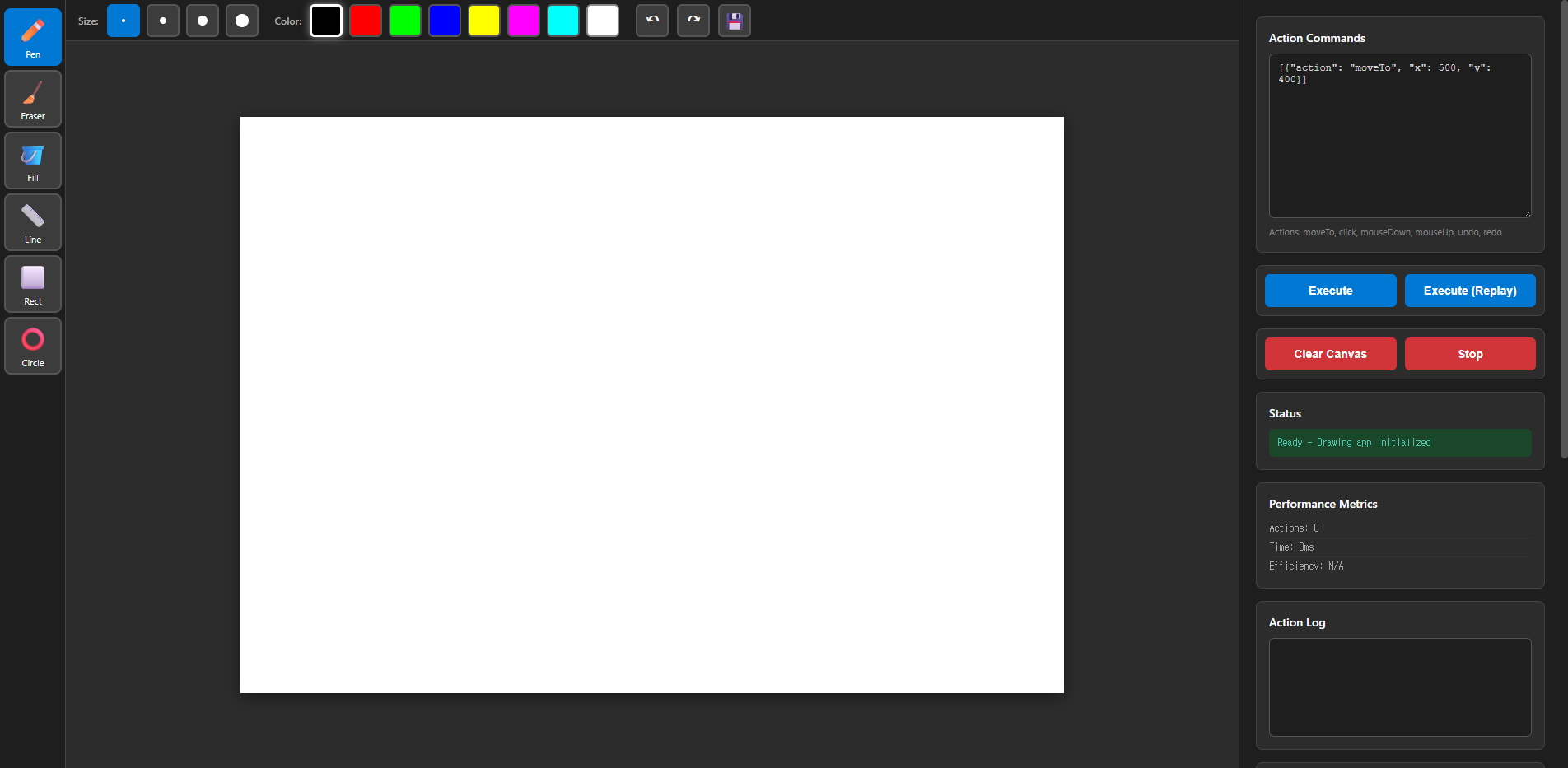}
\caption{DrawingBench UI: toolbar (left), menu (top), 1000$\times$700px canvas (center), control panel (right).}
\label{fig:ui_screenshot}
\end{figure*}

\subsection{UI Layout and Coordinates}

The DrawingBench application uses a browser-based drawing interface with the following layout (see Figure~\ref{fig:ui_screenshot}):

\textbf{Canvas Specifications:}
\begin{itemize}
    \item Size: 1000$	imes$700 pixels
    \item Position: (90, 70) to (1090, 770) in screen coordinates
    \item Background: White (\#FFFFFF)
    \item Border: 1px solid black
\end{itemize}

\textbf{Tool Panel (Left Side):}
\begin{table}[h]
\centering
\caption{Tool Coordinates and Functions}
\small
\begin{tabular}{llll}
\toprule
\textbf{Tool} & \textbf{Coordinates} & \textbf{Size} & \textbf{Function} \\
\midrule
Pen & (35, 45) & 30$	imes$30 px & Free-form drawing \\
Eraser & (35, 125) & 30$	imes$30 px & Erase content \\
Fill & (35, 205) & 30$	imes$30 px & Fill enclosed areas \\
Line & (35, 285) & 30$	imes$30 px & Draw straight lines \\
Rectangle & (35, 365) & 30$	imes$30 px & Draw rectangles \\
Circle & (35, 445) & 30$	imes$30 px & Draw circles \\
\bottomrule
\end{tabular}
\end{table}

\textbf{Color Palette (Top):}
\begin{table}[h]
\centering
\caption{Color Coordinates and Hex Values}
\small
\begin{tabular}{lll}
\toprule
\textbf{Color} & \textbf{Coordinates} & \textbf{Hex Code} \\
\midrule
Black & (405, 25) & \#000000 \\
Red & (429, 25) & \#FF0000 \\
Green & (453, 25) & \#00FF00 \\
Blue & (477, 25) & \#0000FF \\
Yellow & (501, 25) & \#FFFF00 \\
Magenta & (525, 25) & \#FF00FF \\
Cyan & (549, 25) & \#00FFFF \\
White & (573, 25) & \#FFFFFF \\
\bottomrule
\end{tabular}
\end{table}

\textbf{Size Options:}
\begin{itemize}
    \item Small: 2px (default)
    \item Medium: 5px
    \item Large: 10px
\end{itemize}

\section{Detailed Evaluation Criteria}
\label{app:eval_criteria}

Our automated evaluation system checks 8 objective criteria for each drawing submission. Table~\ref{tab:detailed_criteria} provides complete specifications for each criterion.

\begin{table}[h]
\centering
\caption{Complete Evaluation Criteria Specifications}
\label{tab:detailed_criteria}
\small
\setlength{\tabcolsep}{3pt}
\begin{tabular}{p{2.2cm}p{4cm}p{1.2cm}}
\toprule
\textbf{Criterion} & \textbf{Description} & \textbf{Weight} \\
\midrule
Required Tools & All tools specified in task must be selected at least once & 0.20 \\
\midrule
Required Colors & All colors specified must be used in drawing actions & 0.20 \\
\midrule
Minimum Segments & Drawing must contain at least N mouse-drag segments (task-specific) & 0.15 \\
\midrule
Canvas Coverage & Drawing must cover at least X\% of canvas area (typically 0.30) & 0.10 \\
\midrule
Position Constraint & Drawing must satisfy position requirement (center, corner, inside, etc.) & 0.15 \\
\midrule
Size Constraint & If specified, drawn elements must meet size requirements & 0.10 \\
\midrule
Syntax Validity & All action commands must be valid JSON with correct structure & 0.05 \\
\midrule
Coordinate Bounds & All coordinates must be within canvas boundaries (0-1000, 0-700) & 0.05 \\
\bottomrule
\end{tabular}
\end{table}

\subsection{Scoring Formula}

The final score is computed as:
\begin{equation}
\text{Score} = \frac{\sum_{i=1}^{8} w_i \cdot c_i}{\sum_{i=1}^{8} w_i}
\end{equation}
where $w_i$ is the weight and $c_i \in \{0, 1\}$ indicates whether criterion $i$ is satisfied.

\subsection{Error Classification}

Errors are categorized into four types:
\begin{itemize}
    \item \textbf{SYNTAX\_ERROR}: Invalid JSON format, unknown actions, or malformed commands
    \item \textbf{COORDINATE\_ERROR}: Coordinates outside canvas bounds or invalid numeric values
    \item \textbf{LOGIC\_ERROR}: Missing required tools/colors, insufficient segments, or unmet constraints
    \item \textbf{EFFICIENCY\_WARNING}: Excessive actions, redundant tool switches, or suboptimal coverage
\end{itemize}

\section{Task Distribution and Statistics}
\label{app:task_stats}

\subsection{Complete Category Breakdown}

Table~\ref{tab:category_breakdown} shows the distribution of all 250 drawing tasks across 20 categories.

\begin{table}[h]
\centering
\caption{Complete Task Distribution by Category and Difficulty}
\label{tab:category_breakdown}
\scriptsize
\setlength{\tabcolsep}{2pt}
\begin{tabular}{lrrrrrr}
\toprule
\textbf{Category} & \textbf{Easy} & \textbf{Medium} & \textbf{Hard} & \textbf{V.Hard} & \textbf{Total} & \textbf{\%} \\
\midrule
spatial\_reasoning & 35 & 24 & 8 & 4 & 71 & 28.4\% \\
basic\_shapes & 20 & 15 & 5 & 2 & 42 & 16.8\% \\
objects & 12 & 10 & 4 & 2 & 28 & 11.2\% \\
compositional & 4 & 4 & 2 & 1 & 11 & 4.4\% \\
patterns & 3 & 3 & 2 & 1 & 9 & 3.6\% \\
multi\_color & 4 & 2 & 1 & 1 & 8 & 3.2\% \\
position & 3 & 2 & 1 & 1 & 7 & 2.8\% \\
size & 2 & 2 & 1 & 1 & 6 & 2.4\% \\
tool\_usage & 2 & 2 & 1 & 0 & 5 & 2.0\% \\
efficiency\_test & 1 & 1 & 0 & 0 & 2 & 0.8\% \\
creative & 1 & 1 & 0 & 0 & 2 & 0.8\% \\
angle & 1 & 1 & 0 & 0 & 2 & 0.8\% \\
complex & 1 & 1 & 0 & 0 & 2 & 0.8\% \\
scenes & 0 & 0 & 0 & 1 & 1 & 0.4\% \\
precision\_test & 0 & 1 & 0 & 0 & 1 & 0.4\% \\
tool\_switching & 0 & 1 & 0 & 0 & 1 & 0.4\% \\
spatial & 0 & 1 & 0 & 0 & 1 & 0.4\% \\
symmetry & 0 & 0 & 0 & 1 & 1 & 0.4\% \\
texture & 0 & 0 & 1 & 0 & 1 & 0.4\% \\
grid & 0 & 0 & 0 & 1 & 1 & 0.4\% \\
\midrule
\textbf{Total} & \textbf{112} & \textbf{97} & \textbf{21} & \textbf{20} & \textbf{250} & \textbf{100\%} \\
\bottomrule
\end{tabular}
\end{table}

\subsection{Task Complexity Characteristics}

\textbf{Easy Tasks (112):} Single-object drawings with clear specifications (e.g., "Draw a red circle in the center"). Average 3-5 required actions.

\textbf{Medium Tasks (97):} Multi-object scenes or drawings with multiple constraints (e.g., "Draw a blue square above a red circle"). Average 8-12 actions.

\textbf{Hard Tasks (21):} Complex compositions with precise positioning requirements (e.g., "Draw 4 squares in each corner of the canvas"). Average 15-20 actions.

\textbf{Very Hard Tasks (20):} Highly complex patterns requiring extensive procedural accuracy (e.g., "Draw an 8$\times$8 checkerboard pattern"). Average 25+ actions.

\section{Action Sequence Examples}
\label{app:actions}

\subsection{Example 1: Drawing a Red Circle}

{\small
\begin{verbatim}
[
  {"action": "moveTo", "x": 35, "y": 45},
  {"action": "click"},
  {"action": "moveTo", "x": 429, "y": 25},
  {"action": "click"},
  {"action": "moveTo", "x": 590, "y": 420},
  {"action": "mouseDown"},
  {"action": "moveTo", "x": 640, "y": 420},
  {"action": "moveTo", "x": 665, "y": 445},
  {"action": "moveTo", "x": 665, "y": 495},
  {"action": "moveTo", "x": 640, "y": 520},
  {"action": "moveTo", "x": 590, "y": 520},
  {"action": "moveTo", "x": 565, "y": 495},
  {"action": "moveTo", "x": 565, "y": 445},
  {"action": "moveTo", "x": 590, "y": 420},
  {"action": "mouseUp"}
]
\end{verbatim}
}

\subsection{Example 2: Drawing a Blue Rectangle}

{\small
\begin{verbatim}
[
  {"action": "moveTo", "x": 35, "y": 365},
  {"action": "click"},
  {"action": "moveTo", "x": 477, "y": 25},
  {"action": "click"},
  {"action": "moveTo", "x": 400, "y": 300},
  {"action": "mouseDown"},
  {"action": "moveTo", "x": 700, "y": 500},
  {"action": "mouseUp"}
]
\end{verbatim}
}

\subsection{Example 3: Drawing 4 Squares in Corners}

{\small
\begin{verbatim}
[
  {"action": "moveTo", "x": 35, "y": 365},
  {"action": "click"},
  {"action": "moveTo", "x": 120, "y": 100},
  {"action": "mouseDown"},
  {"action": "moveTo", "x": 220, "y": 200},
  {"action": "mouseUp"},
  {"action": "moveTo", "x": 870, "y": 100},
  {"action": "mouseDown"},
  {"action": "moveTo", "x": 970, "y": 200},
  {"action": "mouseUp"},
  {"action": "moveTo", "x": 120, "y": 640},
  {"action": "mouseDown"},
  {"action": "moveTo", "x": 220, "y": 740},
  {"action": "mouseUp"},
  {"action": "moveTo", "x": 870, "y": 640},
  {"action": "mouseDown"},
  {"action": "moveTo", "x": 970, "y": 740},
  {"action": "mouseUp"}
]
\end{verbatim}
}

\subsection{Example 4: Multi-Color Drawing (Red Circle Above Blue Square)}

{\small
\begin{verbatim}
[
  {"action": "moveTo", "x": 35, "y": 445},
  {"action": "click"},
  {"action": "moveTo", "x": 429, "y": 25},
  {"action": "click"},
  {"action": "moveTo", "x": 590, "y": 250},
  {"action": "mouseDown"},
  {"action": "moveTo", "x": 640, "y": 250},
  {"action": "moveTo", "x": 665, "y": 275},
  {"action": "moveTo", "x": 665, "y": 325},
  {"action": "moveTo", "x": 640, "y": 350},
  {"action": "moveTo", "x": 590, "y": 350},
  {"action": "moveTo", "x": 565, "y": 325},
  {"action": "moveTo", "x": 565, "y": 275},
  {"action": "moveTo", "x": 590, "y": 250},
  {"action": "mouseUp"},
  {"action": "moveTo", "x": 35, "y": 365},
  {"action": "click"},
  {"action": "moveTo", "x": 477, "y": 25},
  {"action": "click"},
  {"action": "moveTo", "x": 515, "y": 470},
  {"action": "mouseDown"},
  {"action": "moveTo", "x": 665, "y": 620},
  {"action": "mouseUp"}
]
\end{verbatim}
}

\subsection{Example 5: Complex Pattern (Grid of 6 Circles)}

{\small
\begin{verbatim}
[
  {"action": "moveTo", "x": 35, "y": 445},
  {"action": "click"},
  {"action": "moveTo", "x": 200, "y": 200},
  {"action": "mouseDown"},
  {"action": "moveTo", "x": 280, "y": 280},
  {"action": "mouseUp"},
  {"action": "moveTo", "x": 450, "y": 200},
  {"action": "mouseDown"},
  {"action": "moveTo", "x": 530, "y": 280},
  {"action": "mouseUp"},
  {"action": "moveTo", "x": 700, "y": 200},
  {"action": "mouseDown"},
  {"action": "moveTo", "x": 780, "y": 280},
  {"action": "mouseUp"},
  {"action": "moveTo", "x": 200, "y": 460},
  {"action": "mouseDown"},
  {"action": "moveTo", "x": 280, "y": 540},
  {"action": "mouseUp"},
  {"action": "moveTo", "x": 450, "y": 460},
  {"action": "mouseDown"},
  {"action": "moveTo", "x": 530, "y": 540},
  {"action": "mouseUp"},
  {"action": "moveTo", "x": 700, "y": 460},
  {"action": "mouseDown"},
  {"action": "moveTo", "x": 780, "y": 540},
  {"action": "mouseUp"}
]
\end{verbatim}
}

\subsection{Example 6: Using Fill Tool (House with Filled Roof)}

{\small
\begin{verbatim}
[
  {"action": "moveTo", "x": 35, "y": 45},
  {"action": "click"},
  {"action": "moveTo", "x": 400, "y": 300},
  {"action": "mouseDown"},
  {"action": "moveTo", "x": 400, "y": 550},
  {"action": "moveTo", "x": 650, "y": 550},
  {"action": "moveTo", "x": 650, "y": 300},
  {"action": "moveTo", "x": 400, "y": 300},
  {"action": "mouseUp"},
  {"action": "moveTo", "x": 525, "y": 180},
  {"action": "mouseDown"},
  {"action": "moveTo", "x": 400, "y": 300},
  {"action": "moveTo", "x": 650, "y": 300},
  {"action": "moveTo", "x": 525, "y": 180},
  {"action": "mouseUp"},
  {"action": "moveTo", "x": 35, "y": 205},
  {"action": "click"},
  {"action": "moveTo", "x": 429, "y": 25},
  {"action": "click"},
  {"action": "moveTo", "x": 525, "y": 240},
  {"action": "click"}
]
\end{verbatim}
}

\section{Feedback Examples}
\label{app:feedback}

\subsection{Example 1: Missing Tool Error}

{\scriptsize
\begin{verbatim}
Score: 0.75/1.00 - Good attempt! 1 error(s) found.

ERRORS:
  1. Required tool 'rectangle' was not used.
     Select it by clicking at coordinates (35, 365).

SUGGESTIONS:
  1. Use the rectangle tool for more efficient shape creation.
  2. Drawing coverage is good (0.42).

MISSING CRITERIA:
  - required_tools: ['rectangle']
\end{verbatim}
}

\subsection{Example 2: Insufficient Coverage}

{\scriptsize
\begin{lstlisting}[basicstyle=\scriptsize\ttfamily, breaklines=true]
Score: 0.80/1.00 - Good job! 1 warning(s) found.

WARNINGS:
  1. Drawing is too small (coverage: 0.12).
     Consider using larger coordinates to cover more canvas area.

SUGGESTIONS:
  1. Aim for at least 0.30 coverage of the canvas.
  2. All required tools and colors were correctly used.

CURRENT STATS:
  - Tools used: ['pen']
  - Colors used: ['#FF0000']
  - Segments: 3
  - Coverage: 0.12
\end{lstlisting}
}

\subsection{Example 3: Perfect Score}

{\small
\begin{verbatim}
Score: 1.00/1.00 - Excellent! Perfect score!

All criteria met:
  + Required tools: ['pen']
  + Required colors: ['\#FF0000']
  + Minimum segments: 1
  + Minimum coverage: 0.30
  + Position constraint: center
  + No errors detected

Great job! No improvements needed.
\end{verbatim}
}

\section{Additional Performance Analysis}
\label{app:additional}

\subsection{Model-Specific Performance}

\begin{table}[h]
\centering
\caption{Per-Model Performance Breakdown}
\small
\setlength{\tabcolsep}{4pt}
\begin{tabular}{lrrrr}
\toprule
\textbf{Model} & \textbf{T1 Score} & \textbf{T2 Score} & \textbf{Perfect \%} & \textbf{Time (min)} \\
\midrule
Claude-4 Sonnet & 0.931 & 0.958 & 94.4\% & 84.2 \\
GPT-4.1 & 0.927 & 0.955 & 93.2\% & 86.3 \\
GPT-4.1-mini & 0.919 & 0.949 & 90.8\% & 77.1 \\
Gemini-2.5 Flash & 0.923 & 0.951 & 92.0\% & 90.6 \\
\midrule
\textbf{Average} & \textbf{0.925} & \textbf{0.953} & \textbf{92.6\%} & \textbf{84.6} \\
\bottomrule
\end{tabular}
\end{table}

\subsection{Category-Level Detailed Results}

\begin{table}[h]
\centering
\caption{Complete Category Performance (All 20 Categories)}
\tiny
\setlength{\tabcolsep}{2pt}
\begin{tabular}{lrrrrrr}
\toprule
\textbf{Category} & \textbf{Tests} & \textbf{T1 Avg} & \textbf{T2 Avg} & \textbf{Improve} & \textbf{Perfect T1} & \textbf{Perfect T2} \\
\midrule
spatial\_reasoning & 71 & 0.959 & 0.973 & +0.015 & 53 & 67 \\
basic\_shapes & 42 & 0.945 & 0.968 & +0.023 & 35 & 40 \\
objects & 28 & 0.912 & 0.941 & +0.029 & 20 & 25 \\
compositional & 11 & 0.957 & 0.998 & +0.041 & 9 & 11 \\
patterns & 9 & 0.889 & 0.925 & +0.036 & 6 & 8 \\
multi\_color & 8 & 0.931 & 0.956 & +0.025 & 6 & 7 \\
position & 7 & 0.943 & 0.971 & +0.028 & 5 & 6 \\
size & 6 & 0.917 & 0.950 & +0.033 & 4 & 5 \\
tool\_usage & 5 & 0.960 & 0.980 & +0.020 & 4 & 5 \\
efficiency\_test & 2 & 1.000 & 1.000 & 0.000 & 2 & 2 \\
creative & 2 & 0.924 & 1.000 & +0.076 & 1 & 2 \\
angle & 2 & 0.881 & 0.980 & +0.100 & 1 & 2 \\
complex & 2 & 0.900 & 0.975 & +0.075 & 1 & 2 \\
scenes & 1 & 0.672 & 1.000 & +0.328 & 0 & 1 \\
precision\_test & 1 & 1.000 & 1.000 & 0.000 & 1 & 1 \\
tool\_switching & 1 & 1.000 & 1.000 & 0.000 & 1 & 1 \\
spatial & 1 & 1.000 & 1.000 & 0.000 & 1 & 1 \\
symmetry & 1 & 0.950 & 0.950 & 0.000 & 0 & 0 \\
texture & 1 & 0.875 & 0.925 & +0.050 & 0 & 0 \\
grid & 1 & 0.800 & 0.875 & +0.075 & 0 & 0 \\
\bottomrule
\end{tabular}
\end{table}

\subsection{Error Type Distribution by Difficulty}

\begin{table}[h]
\centering
\caption{Error Distribution Across Difficulty Levels (Turn 1)}
\small
\setlength{\tabcolsep}{4pt}
\begin{tabular}{lrrrr}
\toprule
\textbf{Error Type} & \textbf{Easy} & \textbf{Medium} & \textbf{Hard} & \textbf{Very Hard} \\
\midrule
SYNTAX\_ERROR & 1 & 1 & 0 & 1 \\
COORDINATE\_ERROR & 2 & 3 & 1 & 2 \\
LOGIC\_ERROR & 4 & 5 & 2 & 4 \\
EFFICIENCY\_WARNING & 12 & 15 & 6 & 9 \\
\midrule
\textbf{Total} & \textbf{19} & \textbf{24} & \textbf{9} & \textbf{16} \\
\bottomrule
\end{tabular}
\end{table}

\subsection{Token Usage Statistics}

\begin{table}[h]
\centering
\caption{Average Token Usage per Test}
\small
\begin{tabular}{lrrr}
\toprule
\textbf{Model} & \textbf{Input Tokens} & \textbf{Output Tokens} & \textbf{Total} \\
\midrule
Claude-4 Sonnet & 1,243 & 387 & 1,630 \\
GPT-4.1 & 1,256 & 412 & 1,668 \\
GPT-4.1-mini & 1,198 & 356 & 1,554 \\
Gemini-2.5 Flash & 1,287 & 398 & 1,685 \\
\midrule
\textbf{Average} & \textbf{1,246} & \textbf{388} & \textbf{1,634} \\
\bottomrule
\end{tabular}
\end{table}

\section{Sample Drawing Prompts by Category}
\label{app:prompts}

\subsection{Easy Prompts}
\begin{enumerate}
    \item Draw a red circle in the center of the canvas
    \item Draw a blue square in the top-left corner
    \item Draw a green horizontal line across the middle
    \item Draw a yellow triangle pointing upward
    \item Fill the entire canvas with cyan color
\end{enumerate}

\subsection{Medium Prompts}
\begin{enumerate}
    \item Draw a house with a triangular roof and rectangular body
    \item Draw a simple face with two eyes, a nose, and a mouth
    \item Draw three overlapping circles of different colors
    \item Draw a tree with a brown trunk and green leaves
    \item Create a pattern with alternating red and blue vertical stripes
\end{enumerate}

\subsection{Hard Prompts}
\begin{enumerate}
    \item Draw 4 identical squares, one in each corner of the canvas
    \item Create a 3$	imes$3 grid of circles, all the same size
    \item Draw a star with 5 points using only straight lines
    \item Create a rainbow with 7 colored arcs
    \item Draw a flower with 8 petals arranged symmetrically
\end{enumerate}

\subsection{Very Hard Prompts}
\begin{enumerate}
    \item Draw a checkerboard pattern with 8$	imes$8 squares
    \item Create a spiral pattern starting from the center
    \item Draw a complex geometric mandala with rotational symmetry
    \item Create a pixel art character using a 16$	imes$16 grid
    \item Draw a detailed landscape with hills, trees, sun, and clouds
\end{enumerate}

\section{Evaluation Criteria Details}
\label{app:criteria}

\subsection{Required Tools Criterion}

This criterion checks whether the LLM selected and used specific drawing tools as required by the task.

\textbf{Evaluation Method:}
\begin{enumerate}
    \item Parse action sequence for tool selection actions
    \item Identify clicked coordinates matching tool positions
    \item Verify required tools are in the set of used tools
    \item Score: 1.0 if all required tools used, 0.0 otherwise
\end{enumerate}

\textbf{Common Failure Modes:}
\begin{itemize}
    \item Forgetting to select the tool before drawing
    \item Using wrong tool (e.g., pen instead of rectangle)
    \item Tool state not properly tracked across actions
\end{itemize}

\subsection{Canvas Coverage Criterion}

This criterion measures what percentage of the canvas area is covered by the drawing.

\textbf{Calculation:}
\begin{equation}
\text{Coverage} = \frac{\text{Bounding Box Area}}{\text{Canvas Area}}
\end{equation}

where Bounding Box is the smallest rectangle containing all drawn content.

\textbf{Thresholds:}
\begin{itemize}
    \item Minimum coverage: typically 0.20-0.30
    \item Good coverage: 0.40-0.60
    \item High coverage: >0.60
\end{itemize}

\subsection{Position Constraint Criterion}

This criterion verifies whether drawn elements are positioned correctly according to the prompt.

\textbf{Supported Constraints:}
\begin{itemize}
    \item \texttt{center}: element must be within central 20\% of canvas
    \item \texttt{top-left}: element in top-left quadrant
    \item \texttt{top-right}: element in top-right quadrant
    \item \texttt{bottom-left}: element in bottom-left quadrant
    \item \texttt{bottom-right}: element in bottom-right quadrant
    \item \texttt{corners}: elements in all four corners
\end{itemize}

\section{Implementation Details}
\label{app:implementation}

\subsection{Technology Stack}

\textbf{Frontend:}
\begin{itemize}
    \item HTML5 Canvas API for drawing
    \item JavaScript for UI interactions
    \item Puppeteer for browser automation
\end{itemize}

\textbf{Backend:}
\begin{itemize}
    \item Python 3.11+ for evaluation engine
    \item OpenAI, Anthropic, Google AI APIs for LLM access
    \item JSON for data serialization
\end{itemize}

\textbf{Evaluation Pipeline:}
\begin{enumerate}
    \item Load prompt from dataset
    \item Send to LLM with UI specification
    \item Parse LLM response (JSON action sequence)
    \item Execute actions in browser via Puppeteer
    \item Capture canvas state
    \item Run rule-based evaluation
    \item Generate structured feedback
    \item (Optional) Send feedback to LLM for Turn 2
    \item Store results and metrics
\end{enumerate}

\subsection{Code Availability}

All code, data, and documentation for DrawingBench are available at:

\texttt{[Repository URL to be added upon publication]}

The repository includes:
\begin{itemize}
    \item Complete dataset (250 prompts with metadata)
    \item Drawing application (HTML/JS)
    \item Evaluation engine (Python)
    \item Feedback generator (Python)
    \item LLM testing scripts
    \item Result analysis notebooks
    \item Documentation and examples
\end{itemize}

\section{Limitations and Threats to Validity}
\label{app:limitations}

Several limitations of our study warrant discussion:

\paragraph{Rule-based Evaluation Constraints.} Our automated evaluation system excels at objective criteria (e.g., position, color, coverage) but cannot assess subjective qualities like aesthetic appeal or creative interpretation. Some drawings that violate our criteria might still be considered acceptable or even preferable by human judges. Incorporating human evaluation or learned quality metrics could provide a more holistic assessment.

\paragraph{Text-only Paradigm.} While we demonstrate that LLMs can perform spatial tasks without vision, real-world agents increasingly use multimodal models with visual perception. Our benchmark's text-only design may not fully represent the capabilities or challenges of vision-enabled LLM agents. Extending DrawingBench to a vision-based mode (where models receive canvas screenshots and generate actions) would be a valuable direction for future work.

\paragraph{Limited Task Diversity.} Although our 250 prompts span 20 categories and 4 difficulty levels, they focus primarily on geometric and compositional tasks. Other spatial reasoning aspects, such as 3D spatial reasoning, dynamic movement, or physics-based interactions, are not covered. Expanding the benchmark to include such tasks would broaden its applicability.

\paragraph{Model Selection.} We evaluated four state-of-the-art models, but the LLM landscape is rapidly evolving. Results may vary with newer models, different prompting strategies, or fine-tuned versions specialized for agent tasks. Continuous evaluation of emerging models will be necessary to track progress.

\paragraph{Generalization to Real-world UIs.} Our drawing application is a simplified, controlled environment. Real-world UIs have greater complexity, including dynamic elements, latency, error states, and non-deterministic behaviors. While our benchmark tests core spatial reasoning and action generation, generalizing these findings to production agent systems requires further research.

\section{Broader Impact and Ethical Considerations}
\label{app:ethics}

The development of LLM-based agents capable of UI interaction raises important ethical and societal considerations. On the positive side, such agents could democratize access to digital tools, assist users with disabilities, and automate tedious tasks. However, they also pose risks, including:

\begin{itemize}[noitemsep,topsep=0pt]
    \item \textbf{Automation of harmful activities}: Agents with UI control could be misused for automated phishing, spam, or unauthorized access.
    \item \textbf{Job displacement}: Widespread deployment of UI automation agents might disrupt employment in fields like data entry, software testing, or customer support.
    \item \textbf{Privacy and security}: Agents interacting with UIs may access sensitive information, requiring robust safeguards.
    \item \textbf{Bias and fairness}: If agents are trained or evaluated on biased datasets, they may perpetuate inequities in access or outcomes.
\end{itemize}

Our benchmark, being open-source and focused on a simplified drawing task, has limited direct ethical implications. However, as the techniques and insights from DrawingBench are applied to real-world systems, developers must consider these broader impacts and implement appropriate safeguards.

\section{Future Directions}
\label{app:future}

Building on our findings, we identify several promising directions for future research:

\paragraph{Vision-based Evaluation.} Extending DrawingBench to support vision-language models (VLMs) would enable direct comparison between text-only and vision-augmented spatial reasoning. This could reveal whether visual perception provides significant advantages for spatial tasks or if text-based reasoning suffices.

\paragraph{Expanded Task Coverage.} Incorporating tasks that require temporal reasoning (e.g., animated drawings), 3D spatial concepts (e.g., perspective drawing), or physical simulation (e.g., balancing objects) would broaden the benchmark's scope and challenge models in new ways.

\paragraph{Human-in-the-loop Feedback.} Replacing or augmenting our rule-based feedback with human annotations could test whether models can learn from more naturalistic, less structured guidance. This would also allow exploration of how models handle ambiguous or conflicting feedback.

\paragraph{Fine-tuning and Specialization.} Our study evaluates general-purpose LLMs. Training models specifically on drawing or spatial tasks could reveal the extent to which spatial reasoning can be improved through targeted learning.

\paragraph{Cross-domain Transfer.} Investigating whether strong performance on DrawingBench correlates with success on other agent tasks (e.g., web navigation, code generation) would shed light on the generalizability of spatial reasoning skills.